\documentclass[letterpaper, 10 pt, conference]{ieeeconf}
\IEEEoverridecommandlockouts

\usepackage[colorlinks]{hyperref}
\hypersetup{
    colorlinks,
    linkcolor=black,
    citecolor=black,
    filecolor=magenta,
    urlcolor=cyan,
}

\usepackage{resizegather}

\usepackage{cite}
\usepackage{amsmath,amssymb,amsfonts,mathrsfs}
\usepackage{graphicx}
\usepackage{textcomp}
\usepackage{xcolor}
\usepackage{tcolorbox}
\usepackage{textcomp}
\usepackage{tablefootnote}
\usepackage{threeparttable}
\usepackage{caption, subcaption}
\usepackage{bbm}
\usepackage{dsfont}
\usepackage{tikz}
\usepackage{graphicx}
\usepackage{tkz-euclide}
\usepackage{algorithm}
\usepackage{algpseudocode}
\usepackage{enumerate}
\usepackage{mathtools}
\usepackage{balance}
\usetikzlibrary{positioning}
\usetikzlibrary{shapes}
\usetikzlibrary{shapes.misc}
\usetikzlibrary{shapes.geometric}
\usetikzlibrary{plotmarks}
\usetikzlibrary{intersections}
\usetikzlibrary{calc}
\usetikzlibrary{fit}
\usetikzlibrary{patterns,tikzmark}
\usetikzlibrary{matrix,decorations.pathreplacing,calc}

\tikzset{cross/.style={cross out, draw, 
         minimum size=2*(#1-\pgflinewidth), 
         inner sep=0pt, outer sep=0pt}}

\newcommand{\state}[0]{x}

\newcommand{\vertiii}[1]{{\left\vert\kern-0.25ex\left\vert\kern-0.25ex\left\vert #1 
    \right\vert\kern-0.25ex\right\vert\kern-0.25ex\right\vert}}

\newtheorem{theorem}{Theorem}

\newtheorem{lemma}{Lemma}

\newtheorem{definition}{Definition}

\makeatletter
\renewcommand{\fps@figure}{htp}
\renewcommand{\fps@table}{htp}
\makeatother

\def\BibTeX{{\rm B\kern-.05em{\sc i\kern-.025em b}\kern-.08em
    T\kern-.1667em\lower.7ex\hbox{E}\kern-.125emX}}

\title{Semi-Supervised Safe Visuomotor Policy Synthesis using Barrier Certificates }

\author{Manan Tayal*, Aditya Singh*, Pushpak Jagtap, and Shishir Kolathaya 
\thanks{
}
\thanks{This work was supported in part by PMRF, ARTPARK and Kotak IISc AI-ML Centre, IISc.}
\thanks{All the authors belong to Cyber-Physical Systems, Indian Institute of Science (IISc), Bengaluru.
{\tt\scriptsize \{manantayal, adityasingh, pushpak, shishirk\}@iisc.ac.in}
.
}%
\thanks{* denotes equal contribution.
}%
}
\usepackage{amssymb}
\usepackage{pifont}
\newcommand{\cmark}{\ding{51}}%
\newcommand{\xmark}{\ding{55}}%

\begin{document}

\maketitle

\begin{abstract}
In modern robotics, addressing the lack of accurate state space information in real-world scenarios has led to a significant focus on utilizing visuomotor observation to provide safety assurances. Although supervised learning methods, such as imitation learning, have demonstrated potential in synthesizing control policies based on visuomotor observations, they require ground truth safety labels for the complete dataset and do not provide formal safety assurances. On the other hand, traditional control-theoretic methods like Control Barrier Functions (CBFs) and Hamilton-Jacobi (HJ) Reachability provide formal safety guarantees but depend on accurate knowledge of system dynamics, which is often unavailable for high-dimensional visuomotor data. To overcome these limitations, we propose a novel approach to synthesize a semi-supervised safe visuomotor policy using barrier certificates that integrate the strengths of model-free supervised learning and model-based control methods. This framework synthesizes a provably safe controller without requiring safety labels for the complete dataset and ensures completeness guarantees for both the barrier certificate and the policy. We validate our approach through distinct case studies: an inverted pendulum system and the obstacle avoidance of an autonomous mobile robot. Simulation videos of both case studies can be viewed on the project webpage\footnote{\label{note: sim videos}\url{https://tayalmanan28.github.io/sspl/}}.

\end{abstract}

\section{Introduction}
\label{section: introduction}

With the swift incorporation of autonomous systems across different domains, ensuring their safety has become essential. However, the absence of accurate state space representations in large-scale practical systems has driven research efforts to investigate the use of visuomotor feedback to manage operations in safety-critical environments. Research has demonstrated that visuomotor data can be effectively integrated into control strategies, allowing robots to autonomously navigate and make decisions. However, for safety-critical systems, the challenge remains in formally verifying the safety of such controllers. Supervised learning approaches like imitation learning \cite{Osa_2018, pan2019agile} help us in synthesizing controllers with visuomotor feedback. However, this approach relies on a large number of near-optimal expert demonstrations, which may not always be readily available. Moreover, they don't provide any formal guarantees on safety, which poses a significant risk to the systems. This risk is particularly enhanced when these policies are applied to out-of-distribution environments where their reliability cannot be ensured. 

Conventional optimal-control methods based on the Hamilton-Jacobi reachability analysis framework \cite{8263977, RA-UAV, 9561949, singh2024imposingexactsafetyspecifications} have been instrumental in addressing safety concerns by framing them as constraints or determining safe control strategies offline. These methods rely on solving partial differential equations (PDEs) to derive a safety value function. However, the computational and memory requirements for solving these PDEs increase exponentially with the size of the system’s state space, making their direct application impractical for large-scale systems.


Another promising approach to addressing safety challenges lies in the use of Control Barrier Functions (CBFs) \cite{ames2014control}, which offer a practical framework for synthesizing safe controllers for control affine systems \cite{Ames_2017}\cite{ames2019control}. CBFs enable the formulation of controllers via Quadratic Programs (QPs), which can be solved at high frequencies using modern optimization techniques. This method has been successfully applied in various safety-critical scenarios, such as adaptive cruise control~\cite{ames2014control}, aerial maneuvers \cite{7525253,tayal2023control}, and legged locomotion~\cite{ames2019control,nguyen2015safety}. Nevertheless, while CBFs have proven their utility, their applicability is constrained to control-affine systems with no bound on control inputs. In contrast, Control Barrier Certificates (CBCs) \cite{1273063,jagtap2020formal} provide a more flexible approach by facilitating the synthesis of safe controllers with input constraints. Furthermore, CBCs are applicable to general nonlinear systems, offering a more scalable solution for ensuring safety.


Unfortunately, these approaches generally rely on known/approximate models. While approximate models are often available in systems with state feedback \cite{jagtap2020control}, the absence of a predictive model for visuomotor observations has posed challenges for applying CBFs in such scenarios. Recent studies have started exploring CBFs for safe control using visuomotor feedback~\cite{10161482, 10160805, 10077790, 10610647, harms2024neuralcontrolbarrierfunctions, kumar2024latentcbf}, but the requirement of control affine systems limits their broader applicability. For instance, a NeRF-based CBF approach in~\cite{10161482} shows promise for visuomotor control but is computationally expensive, making real-time execution impractical. Other methods, such as~\cite{10160805}, employ Generative Adversarial Networks (GANs) to infer 3D obstacle positions and velocities from images to calculate geometric CBFs. Approaches like~\cite{harms2024neuralcontrolbarrierfunctions, kumar2024latentcbf} generate CBFs in a latent space derived from visuomotor data.
However, \cite{kumar2024latentcbf} utilizes a Lipschitz autoencoder with reconstruction loss to encode the latent space, but this alone does not guarantee the clear separation of safe and unsafe regions, a crucial requirement for accurate barrier function learning. Moreover, none of these approaches provides completeness guarantees, meaning that the learned functions only satisfy CBF conditions on the training data and not necessarily across the entire state space.

To overcome these limitations, we introduce a semi-supervised framework for synthesizing provably safe visuomotor policies by jointly learning it with a Control Barrier Certificate (CBC). To summarize, this paper makes the following contributions:
\begin{itemize}
    \item We propose a new training framework that synthesizes a Control Barrier Certificate (CBC) and a safe policy using visuomotor observations. The learned CBC inherently satisfies completeness guarantees, eliminating the need for any post hoc formal verification. 
    \item We introduce a Safety-based Latent Dynamics (SaLaD) model that learns a latent representation where safe and unsafe regions are distinctly separable. This preserves only the essential information for this separation and improves computational efficiency.
    \item To address the issue of non-zero (latent) dynamics loss, we derive a consistency condition to formally verify the barrier certificate. This ensures that the policy meets CBC conditions despite errors in learned dynamics. 
    \item We test the proposed framework on two different control problems: an inverted pendulum system and an autonomous mobile robot for obstacle avoidance, demonstrating that it synthesizes a policy that satisfies all prescribed safety constraints for both systems. 

\end{itemize}

The rest of this paper is organized as follows. Section \ref{section: problem_formulation} presents the preliminaries and formulates the problem. Section \ref{section: sspl} presents the proposed method for jointly synthesizing CBCs and a safe policy using neural networks, including the construction of the loss functions and the corresponding training algorithm. Section \ref{section: Experiments} outlines the case studies. Finally, the conclusions are summarized in Section \ref{section: Conclusions}.

\section{Preliminaries and Problem Formulation} 
\label{section: problem_formulation}

In this section, we will formally introduce Control Barrier Certificates (CBCs), existing work on providing completeness guarantees for Neural CBCs, and finally, the problem formulation for the paper.
\subsection{Notations}
For a set \( A \), we define the indicator function of \( A \), denoted by \( \mathds{1}_A(x) \), where, $ ( \mathds{1}_A(x)) = 1$ when $x \in A$ and 0 otherwise. The complement of a set $A$ within a set $B$ is denoted by $B \backslash A$.

\subsection{System Description}
A discrete-time control system is a tuple $\mathbb{S} = (X, U, f)$, where $X \subseteq \mathbb{R}^n$ is the state set of the system, $U \subseteq \mathbb{R}^{n_u}$ is the input set of the system, and $f: X \times U \to X$ describes the state evolution of the system via the following difference equation:
\begin{equation}
\label{eq: system_eq}
\begin{aligned}
    x(t + 1) = f(x(t), u(t)), \forall t \in \mathbb{N},
\end{aligned}
\end{equation}
where $x(t) \in X$ and $u(t) \in U$, $\forall t \in \mathbb{N}$, denote the state and input of the system, respectively.

Consider a set $\mathcal{C}$ defined as the \textit{sub-zero level set} of a continuous function $B: X\subseteq \mathbb{R}^n \rightarrow \mathbb{R}$ yielding,
\begin{align}
\label{eq:setc1}
	\mathcal{C}                        & = \{ \state \in X \subset \mathbb{R}^n : B(\state) \leq 0\} \\
\label{eq:setc2}
	X-\mathcal{C} & = \{ \state \in X \subset \mathbb{R}^n : B(\state) > 0\}.
\end{align}
We further restrict the class of $\mathcal{C}$ where its interior and boundary are precisely the sets given by $\text{Int}\left(\mathcal{C}\right) = \{ \state \in X \subset \mathbb{R}^n : B(\state) < 0\}$ and $\partial\mathcal{C} = \{ \state \in X \subset \mathbb{R}^n : B(\state) = 0\}$, respectively.

\subsection{Control Barrier Certificates (CBCs)}

In this section, we introduce the notion of a control barrier certificate, which provides sufficient conditions together with controllers for the satisfaction of safety constraints. 

\begin{definition}{
\label{definition: CBC definition}
A function $B : X \to \mathbb{R}^{+}_{0}$ is a control barrier certificate for a discrete-time control system $\mathbb{S} = (X, U, f)$ if for any state $x \in X$ there exists an input $u \in U $, such that
\begin{equation}
\label{eq: cbc_condition}
\begin{aligned}
    \! B(f(x,u)) \! \leq \! B(x),
\end{aligned}
\end{equation}}
\end{definition}

The following lemma allows us to synthesize controllers for discrete-time control system $\mathbb{S}$, ensuring the satisfaction of safety properties. 

\begin{lemma}[\hspace{-0.01em}\cite{jagtap2020compositional}]
    \textit{For a discrete-time control system $\mathbb{S}=(X,U,f)$, safe set $X_s \subseteq \mathcal{C}$, and unsafe set $X_u \subseteq X-\mathcal{C}$, the existence of a control barrier certificate, $B$, as defined in Definition \ref{definition: CBC definition}, under a control policy $\pi:X\rightarrow U$ implies that the sequence state in $\mathbb{S}$ starting from $x_s \in X_s$ under the policy $\pi$ do not reach any unsafe states in $X_u$.}
\end{lemma}

The zero-level set of the CBC $B(x)=0$ separates the unsafe regions from the safe ones. For an initial state $x_0$ within the safe region ($x_0 \in X_s$), $B(x_0) \leq 0$ by condition \eqref{eq:setc1}. According to equation \eqref{eq: cbc_condition}, which ensures $B(x)$ remains non-increasing, the level set is not crossed, preventing access to unsafe regions. Therefore, ensuring system safety requires computing appropriate control barrier certificates and corresponding control policies.

\subsection{Neural CBCs with Completeness Guarantees}
To formally verify CBCs, it is essential that they meet completeness guarantees, meaning the learned functions must fulfill condition \eqref{eq: cbc_condition} not only on the finite training samples but also throughout the entire state space. To ensure this, a validity condition was proposed by \cite{tayal2024learning,anand2023formally} to ensure completeness over the entire state space. Specifically, we introduce a positive scalar margin, $\psi$, to the inequalities \eqref{eq:setc1}, \eqref{eq:setc2} and \eqref{eq: cbc_condition}, as follows:
\begin{equation}
    \begin{aligned}
        q_i(x) +\psi \leq 0, \forall i \in {1,2,3},
    \end{aligned}
\end{equation}
where $q_1(x) = (B(x))\mathds{1}_{X_{s}}$, $ q_2(x) = (-B(x))\mathds{1}_{X_{u}} $ and $ q_3(x) = B(f(x,u)) - B(x)$. The following theorem provides a theoretical lower bound of $\psi$.

\begin{theorem}\label{thm: Completeness Guarantees}\textit{
Consider a discrete-time robotic system and initial safe and unsafe sets $X_{s}, X_{u} \subseteq X$, respectively. Given $\bar{\epsilon}$, suppose we sample  $N$ data points: $x_{i} \in X, i \in\{1, \ldots, N\}$, such that $\left\|x-x_{i}\right\| \leq \bar{\epsilon}$. Let  $B_{\theta}$ be the neural network-based CBC with trainable parameters $\theta$. Then $B_{\theta}$ is a valid CBC over the entire state space $X$, if the following condition holds:}
\begin{equation} \label{eq: completeness_condition}
    \psi^{*} \geq L_{B} \bar{\epsilon},
\end{equation}
\textit{where $L_{B}$ is the Lipschitz constant of $B_{\theta}$.}
\end{theorem}
\begin{proof}
    For any $x$ and any $k\in\{1,2,3\}$, we know that:
      \begin{equation*} 
          \begin{aligned}
          q_k(x) & =  q_k(x) - q_k(x_i) + q_k(x_i)\\
          & \leq L_{B}\left\|x - x_i \right\| - \psi^*\\
          & \leq L_{B}\bar{\epsilon} - \psi^* \leq 0.
          \end{aligned}
      \end{equation*}
       Hence, if $q_{k}(x), k \in\{1,2,3\}$ satisfies above condition, then the $B_{\theta}$ is a valid CBC, satisfying conditions \eqref{eq: cbc_condition}
\end{proof}

\subsection{Problem Formulation}

\textit{Given a discrete-time robotic system $\mathbb{S}$ as defined in \eqref{eq: system_eq}. Let $\mathcal{S}$ represent samples of visuomotor observations from the safe region, $\mathcal{U}$ represent samples from the unsafe region, and $\mathcal{D}$ denote the complete set of samples (both labeled and unlabeled). The objective is to devise an algorithm to jointly synthesize a provably correct parameterised barrier certificate $B_{\theta}$ and a safe parameterised policy $\pi_{\theta}$, with parameter $\theta$ such that it satisfies the condition \eqref{eq: cbc_condition} over the entire state space, using a finite number of samples.}

In the following section, we propose an algorithmic approach to solve the above problem.

\section{Semi-Supervised Policy Learning Framework }
\label{section: sspl}

We introduce a semi-supervised policy learning framework that employs control barrier certificates (CBCs) to jointly learn both the barrier certificate and a provably safe policy. Specifically, this framework leverages the forward invariance properties of CBCs to ensure that the learned policy remains within the safe set, provided the initial conditions are within this safe set. 
However, due to the absence of a barrier certificate, we lack access to the safe set $\mathcal{C}$. Hence, we start with initial safe and unsafe sets $X_s \subseteq \mathcal{C}$ and $X_u \subseteq X-\mathcal{C}$, respectively, such that any trajectory starting in $X_s$ never enters $X_u$, which makes this framework semi-supervised. 
We collect the data sets $\mathcal{S}, \mathcal{U}$ and $\mathcal{D}$ corresponding to $N$ visuomotor data points sampled from the initial safe set $X_s \subseteq \mathcal{C}$, initial unsafe set $X_u \subseteq X-\mathcal{C}$, and state set $X$, respectively. 

We represent the barrier functions and the policy as neural networks, denoted as $B_{\theta}$ and $\pi_{\theta}$ respectively.
These functions are learned in a latent space ($z$), encoded by representation network $E_{\theta}$ and use the learned (latent) dynamics \(d_{\theta}\), given by the SaLaD model (Subsection \eqref{subsection: salad}). For notational simplicity, we denote parameters of all the above mentioned neural networks as $\theta$ (online) and $\theta^-$ (target; slow-moving average of $\theta$) as combined feature vectors.


\subsection{Safety based Latent Dynamics (SaLaD)}\label{subsection: salad}
Separability of safe and unsafe regions in the feature space is crucial for learning an accurate barrier certificate. However, in real space, the boundary separating safe and unsafe regions may be complex, which can pose challenges in accurately learning the barrier function. To mitigate this, we propose \textbf{SaLaD}, a \textbf{Sa}fety-driven \textbf{La}tent \textbf{D}ynamics model that learns a latent representation of the environment where the safe and unsafe regions are sufficiently separated while maintaining the state consistency in the latent space. We achieve this by only modelling the elements of the environment that help in segregating the latent space into safe and unsafe regions rather than attempting to model the environment itself. 

\begin{figure}[t]
\label{fig:salad}
    \centering
    \includegraphics[width=\linewidth]{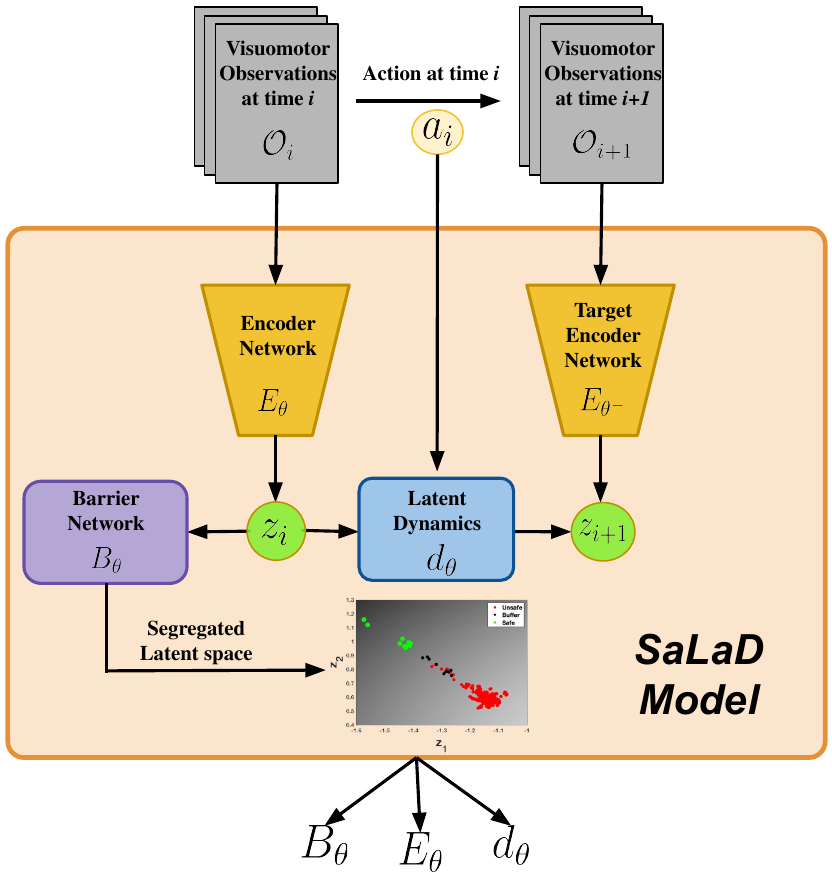}
    
\caption{Presents the architecture of the Safety-driven Latent Dynamics (SaLaD) model. \textbf{Training the SaLaD model} the visuomotor observation $\mathcal{O}_i$
is encoded by $E_{\theta}$ into a latent representation $z_i$. Then, SaLaD
predicts the next latent states $z_{i+1}$ as well as the Barrier Certificate over the safe and unsafe datasets for each latent state, and we optimize SaLaD using Equation \eqref{eq: salad_loss}. Next, visuomotor observation $\mathcal{O}_{i+1}$ is encoded using target net $E_{\theta^-}$ and used as latent targets only during training.}
\end{figure}



Given an observation $\mathcal{O}_i$ observed at time-step $i$, the network $E_{\theta}$ encodes $\mathcal{O}_i$ into a latent representation $z_i$: 
\begin{equation}
    \begin{aligned}
        & z_{i} = E_{\theta}(\mathcal{O}_i);
    \end{aligned}
\end{equation}
From $z_i$ and a random action $a_i$ taken at time-step $i$, SaLaD then predicts the latent dynamics $d_{\theta}$ (latent representation $z_{i+1}$ of the following time-step). Moreover, given the safe and unsafe datasets $\mathcal{S}, \mathcal{U}$ respectively, we learn the barrier certificate $B_{\theta}$, that helps in segregating safe and unsafe regions. 

Now, let us consider the following loss functions satisfying the conditions over the training data sets $\mathcal{S}, \mathcal{U}, \mathcal{D}$ 
as follows:
\begin{equation}
\label{eq: salad_loss}
    \begin{aligned}
        \mathcal{L}_{SaLaD}(\theta)= & \xi_1 \sum_{z_{i} \in \mathcal{S}} \max \left(0,B_{\theta}(z_{i}) + \psi \right)  +\\
        & \xi_2 \sum_{z_{i} \in \mathcal{U}} \max \left(0, -B_{\theta}(z_{i}) + \psi\right) +\\
        & \xi_3 \sum_{z_{i} \in \mathcal{D}} \left( E_{\theta^-}(\mathcal{O}_{i+1}) - d_{\theta}(z_i, a_i)  \right)^2,
    \end{aligned}
\end{equation}
where $d_{\theta}$ represents the latent space dynamics  network parameterized by $\theta$. The first term in loss function $\mathcal{L}_{SaLaD}$ represents the loss for safe states, second term represents the loss for unsafe states, respectively. The third term in $\mathcal{L}_{SaLaD}$ represents the latent state consistency loss. $\xi_1$, $\xi_2$ and $\xi_3$ represents the positive relative weights of each of the losses. $\psi > \psi^*$ is a scalar margin introduced to guarantee completeness of the solution, where $\psi^*$ is given by Eq. \eqref{eq: completeness_condition}.  





\subsection{Formal Verification Integration}
In accordance with Theorem~\eqref{thm: Completeness Guarantees}, it is necessary to establish a bound on the neural network's Lipschitz constant to ensure completeness guarantees. To train neural networks with Lipschitz bounds, we have the following lemma:
\begin{lemma}[\hspace{-0.002em}\cite{pauli2021training}]\label{lemma: lmi_const}
{Suppose $f_{\theta}$ is an $l$-layered feed-forward neural network with $\theta$ as a trainable parameter, then a certificate for $L$-Lipschitz continuity of the neural network is given by the semi-definite constraint $M(\theta, \Lambda) :=$} \\
\begin{equation*}
\begin{aligned}
    \begin{bmatrix} 
    A\\
    B
    \end{bmatrix}^T \begin{bmatrix} 
    2 \alpha\beta\Lambda & -(\alpha + \beta)\Lambda\\
    -(\alpha + \beta)\Lambda & 2\Lambda
    \end{bmatrix}
    \begin{bmatrix} 
    A\\
    B
    \end{bmatrix} \\
    + \begin{bmatrix} 
    L^2 \textbf{\textit{I}} & 0  & 0  & 0\\
    0 & 0 & 0 & 0\\
    0 & 0 & 0 & -\theta_l^T\\
    0 & 0 & -\theta_l &  \textbf{\textit{I}}
    \end{bmatrix} \succeq 0,
\end{aligned}
\end{equation*}
where 
\begin{equation*}
    A = \begin{bmatrix} 
    \theta_0 & \dots & 0  & 0\\
    \vdots & \ddots & \vdots & \vdots\\
    0 & \dots  & \theta_{l-1} & 0
    \end{bmatrix}, 
    B = \begin{bmatrix} 
    0 & I
    \end{bmatrix},
\end{equation*}
{$(\theta_0, \dots \theta_l) $ are the weights of the neural network, $\Lambda \in \mathbb{D}_{+}^{n}$ where $\mathbb{D}_{+}^{n}$ denotes an $n$-dimensional diagonal matrix with positive entries. $\alpha$ and $\beta$ are the minimum and maximum slopes of the activation functions, respectively. }
\end{lemma}

By ensuring that the log-det of a function, $M$, is less than 0, we guarantee the satisfaction of the above semi-definite constraint.
The loss functions characterizing the satisfaction of Lipschitz bound is given as:
\begin{equation}\label{eq: lmi_loss}
    \begin{aligned}
     \mathcal{L}_{M}&(\theta, \Lambda)= -\log \operatorname{det}(M(\theta, \Lambda)),
    \end{aligned}
\end{equation}
where $M$ is the semi-definite matrix corresponding to the Lipschitz bounds $L_{B}$ and $\Lambda$ is a trainable parameter.

\subsection{Controller Synthesis}

In this subsection, we will discuss the synthesis of a safe policy. Based on Lemma 1, satisfying condition \eqref{eq: cbc_condition} ensures that the synthesized policy prevents trajectories starting in the safe set from entering unsafe regions while also ensuring that trajectories originating in unsafe regions eventually reach the safe set. 
The loss function corresponding to the satisfaction of condition \eqref{eq: cbc_condition} is formulated as follows:

\begin{equation}
\label{eq: syn_loss}
    \begin{aligned}
         \mathcal{L}_{syn}(\theta)= 
        &  \sum_{x_{i} \in \mathcal{D}} \max (0, B_{\theta^-} (d_{\theta}(E_{\theta^-}(\mathcal{O}_{i}), \pi_{\theta}(z_i))) + \eta  \\
        &- B_{\theta}(z_{i}) + \psi  ),
    \end{aligned}
\end{equation}
where $\eta > \eta^* $ is a scalar value introduced to compensate for the error in learned (latent) dynamics. The following theorem provides a theoretical lower bound on the value of $\eta$:

\begin{theorem}\label{thm: Dynamics Guarantees}
\textit{Consider a discrete time robotic system and the (latent) dynamics learned using SaLaD model. Let $\delta$ represent the maximum latent state consistency error, then the condition \eqref{eq: cbc_condition} is valid, if the following condition holds:}
\begin{equation} \label{eq: dyn_condition}
     \eta^{*} \geq L_{B} \delta,
\end{equation}

\textit{where $L_{B}$ is the Lipschitz constant of $B_{\theta}$ and
\begin{equation}
    \delta = \max_{i \in \{1, \dots, N \} } \|d_{\theta}(z_{i}, u) - E_{\theta}(\mathcal{O}_{i+1}) \|_{2}.
\end{equation}}
\end{theorem}

\begin{proof}
\begin{equation*} 
    \begin{aligned}
    z_{i+1} &= E_{\theta}(\mathcal{O}_{i+1}) \\
    B(z_{i+1}) - B (d_{\theta}(z_{i}, u))  & \leq | B(z_{i+1}) - B (d_{\theta}(z_{i}, u)) | \\
    & \leq L_B \|z_{i+1} - d_{\theta}(z_{i}, u)\|_2\\
    & \leq L_B \delta.
    \end{aligned} 
\end{equation*}
Now, if we have a $\eta^*$ such that $L_{B} \delta \leq \eta^{*}$, then, we have,
\begin{equation*} 
    \begin{aligned}
 B(z_{i+1}) -  B (d_{\theta}(z_{i}, u)) & \leq \eta^* \\
 B(z_{i+1}) &\leq B (d_{\theta}(z_{i}, u)) + \eta^* \\
 B(z_{i+1}) - B(z_i) &\leq B (d_{\theta}(z_{i}, u)) + \eta^* - B(z_i)\\
 \max(0, B(z_{i+1}) - B(z_i)) &\leq \\ \max(0, B (d_{\theta}(&z_{i}, u)) + \eta^* - B(z_i))
    \end{aligned}
\end{equation*}
Therefore, choosing an $\eta \geq \eta^*$ will ensure the satisfaction of condition \eqref{eq: cbc_condition}, despite the error in learned (latent) dynamics.

\end{proof}

{\textit{Improving the performance of the synthesized controller:}}
Since CBCs only guarantee safety and not performance, we introduce the following loss term to ensure that the synthesized policy, $\pi_{\theta}$, enhances the safety of the user-defined policy, $\pi_{user}$, while remaining close to the baseline performance. 
\begin{equation}
\label{eq: perf_loss}
    \begin{aligned}
         \mathcal{L}_{\pi}(\theta)= \|\pi_{\theta} - \pi_{user}\|_{2}
    \end{aligned}
\end{equation}

\begin{algorithm}
\caption{Semi-Supervised Safe Policy Synthesis}\label{alg:S3P}
\begin{algorithmic}
\Require Data Sets: $\mathcal{S}, \mathcal{U}, \mathcal{D}$, Lipschitz Bounds: $L_B$
\State Initialise($\theta, \psi, \Lambda$)
\State $\mathcal{O}_i \gets sample(\mathcal{S}, \mathcal{U},\mathcal{D})$
{\For{${i}=1$ to ${iter}_{warmstart}$}
{\\ \quad $\psi \gets L_B \bar\epsilon$ \Comment{From eq. \eqref{eq: completeness_condition}}\\ \quad $\mathcal{L}_{SaLaD} \gets (B_{\theta}, E_{\theta}, d_{\theta}, \mathcal{O}_i, a_i, \psi)$ \Comment{From eq. \eqref{eq: salad_loss}} \\
\quad $\theta \gets Learn(\mathcal{L}_{SaLaD}(\theta), \theta)$ \\ 
  \quad $B_{\theta}, E_{\theta}, d_{\theta} \gets \theta$ \\ \quad $\bar\epsilon \gets (E_{\theta}, \mathcal{D})$ 
}}
\\
\\
$\mathcal{L}_{total}(\theta) = \lambda_1 \mathcal{L}_{syn}(\theta) + \lambda_2 \mathcal{L}_{SaLaD}(\theta) + \lambda_3 \mathcal{L}_{\pi}(\theta)$ \\ \Comment{From eq. \eqref{eq: salad_loss},\eqref{eq: syn_loss},\eqref{eq: perf_loss}}
\While{$\mathcal{L}_{total} > 0$ or $\mathcal{L}_M \not\leq 0$ }
    \State $\psi \gets L_B \bar\epsilon$  \Comment{From eq. \eqref{eq: completeness_condition}}
    \State $\eta \gets L_B \delta$  \Comment{From eq. \eqref{eq: dyn_condition}}
    \State $\mathcal{L}_{total} \gets (B_{\theta}, E_{\theta}, d_{\theta}, \pi_{\theta}, \mathcal{O}_i, \psi, \eta)$     
    \State $\theta \gets Learn(\mathcal{L}_{total}(\theta), \theta)$
    \State $\mathcal{L}_{M} \gets (\theta, \Lambda)$ \Comment{From eq. \eqref{eq: lmi_loss}}
    \State $\theta, \Lambda \gets Learn(\mathcal{L}_{M})$
    \State $B_{\theta}, \pi_{\theta}, E_{\theta}, d_{\theta} \gets \theta$
    \State $\mathcal{D} \gets \pi_{\theta} $ (Rollouts)
    \State $\bar\epsilon \gets (E_{\theta}, \mathcal{D})$
    \State $\delta \gets (E_{\theta}, d_{\theta}, \mathcal{D})$
\EndWhile
\end{algorithmic}
\end{algorithm}

\begin{figure*}[htp]
\begin{subfigure}{.25\textwidth}
    \centering
    \includegraphics[width=\textwidth]{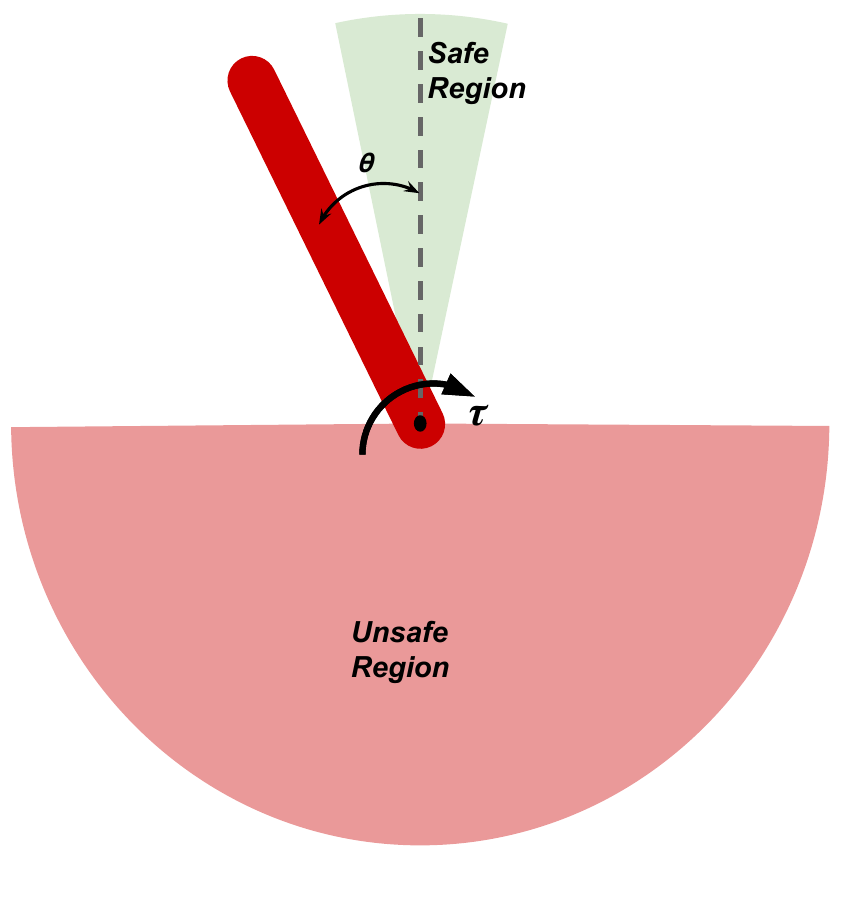}
    \subcaption{}
    \label{fig:sys_pend}
\end{subfigure}
\begin{subfigure}{.37\textwidth}
    \centering
    \includegraphics[width=\textwidth]{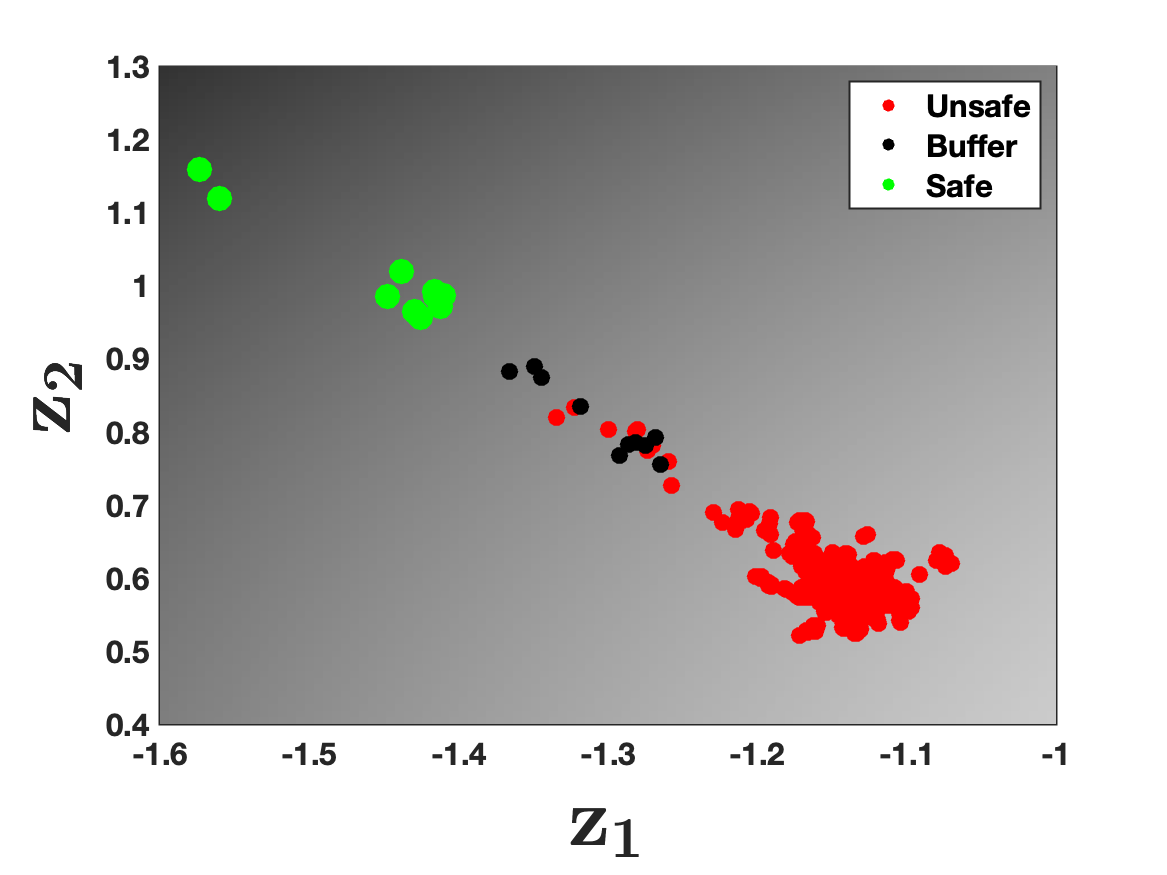}
    \subcaption{}
    \label{fig:plot_latent_pend}
\end{subfigure}
\begin{subfigure}{.37\textwidth}
    \centering
    \includegraphics[width=\textwidth]{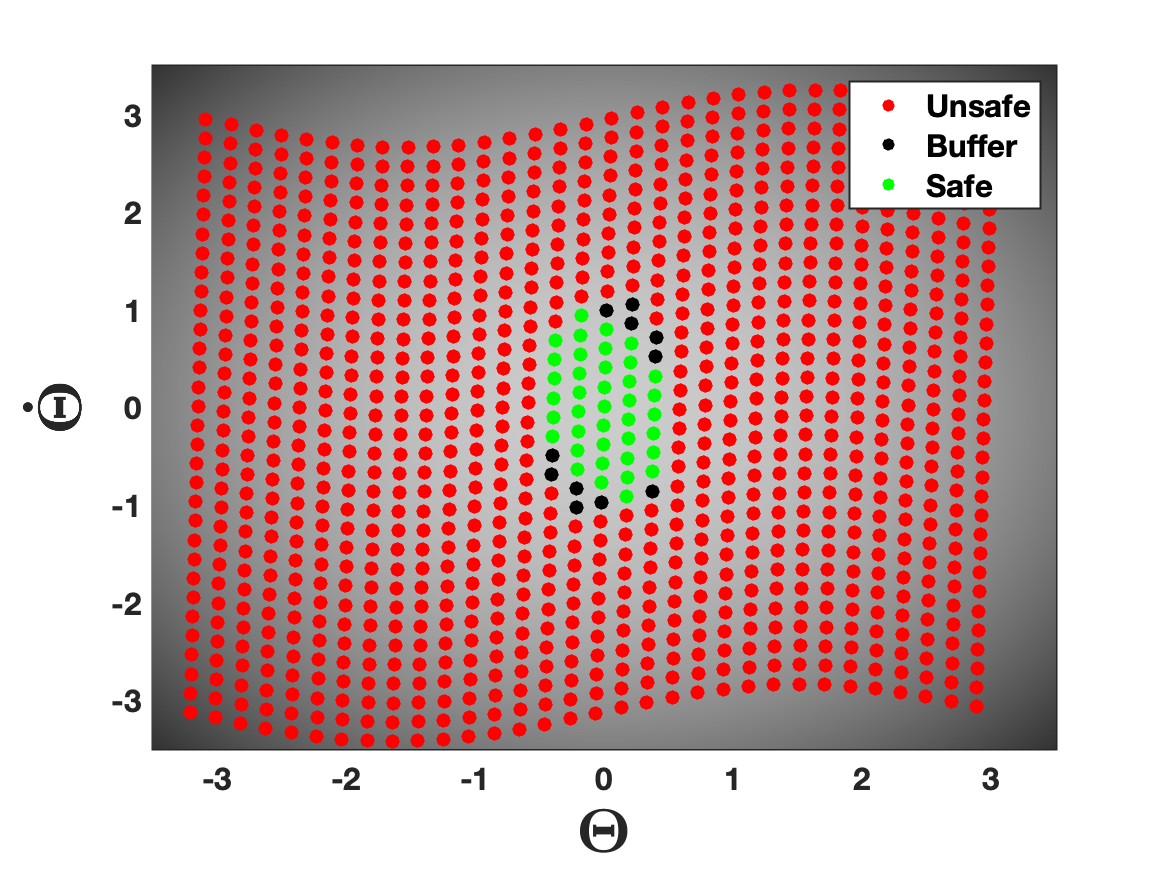}
    \subcaption{}
    \label{fig:plot_real_pend}
\end{subfigure}
\\
\begin{subfigure}{1.0\textwidth}
    \centering
    \includegraphics[width=\textwidth]{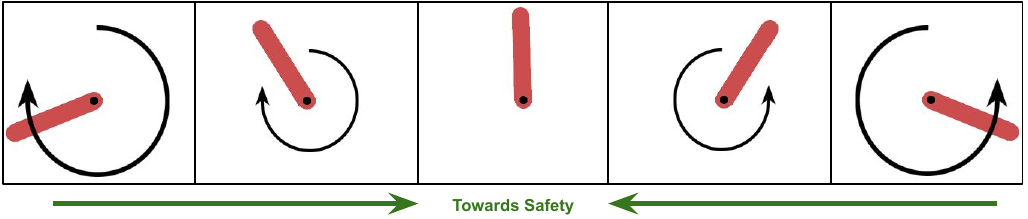}
    \subcaption{}
    \label{fig:sspl_pend}
\end{subfigure}%

\caption{This figure demonstrates the results of our framework applied to OpenAI Gym's Pendulum environment.
\textbf{Fig. \ref{fig:sys_pend}} illustrates the system diagram, where the light green area represents the safe region (used for sampling safe states, $\mathcal{S}$), and the light red area denotes the unsafe region (used for sampling unsafe states, $\mathcal{U}$). \textbf{Fig. \ref{fig:plot_latent_pend}} shows the representation of safe, unsafe and buffer states. The figure shows that the safe and unsafe regions are well segregated. \textbf{Fig. \ref{fig:plot_real_pend}} visualizes the safe, unsafe, and buffer points in the $\Theta-\dot{\Theta}$ plane. 
We observe that the boundary of the trained barrier function (black dots) successfully separates the unsafe and safe regions. 
\textbf{Fig. \ref{fig:sspl_pend}} shows the pendulum’s trajectory generated by the learned policy $\pi_{\theta}$, where the size of the arrows indicates the magnitude of the control actions produced by the policy.
}
\end{figure*}

\subsection{Training Scheme}
The training procedure is divided into the following stages:

\subsubsection{\textbf{Warm Starting the SaLaD Model}} To initialize our training process, we use a warm start approach by optimizing the SaLaD model with the loss function $\mathcal{L}_{SaLaD}$ on samples generated from uniform random actions over a set number of epochs. This method reduces initial model biases, leading to more stable and efficient learning. It also accelerates convergence by offering a well-informed starting point, minimizing the risk of poor local minima, and enhancing overall model performance through more balanced action space exploration.

\subsubsection{\textbf{Iterative Training with Controller Synthesis}} During the training process, we follow an iterative approach consisting of four key steps: 
\begin{enumerate}[i.]
    \item Collecting data through roll-outs from the policy $\pi_{\theta}$.
    \item Improving the policy by optimizing the loss functions $\mathcal{L}_{syn}$ and optionally $\mathcal{L}_{\pi}$.
    \item Fine-tuning the SaLaD model over newly collected data.
    \item Optimizing $\mathcal{L}_{M}$ to ensure that the learned barrier function maintains Lipschitz continuity.
\end{enumerate}
This iterative procedure helps in progressively refining the policy and model, thereby improving overall performance and stability. The overall algorithm is summarised in Algorithm \eqref{alg:S3P}.

\section{Case Studies}
\label{section: Experiments}
In this section, we assess the efficacy of our proposed framework through two distinct case studies: the inverted pendulum system and the obstacle avoidance of an autonomous mobile robot. Both case studies are conducted on a computing platform equipped with an Intel  i9-11900K CPU, 32GB RAM, and NVIDIA GeForce RTX 4090 GPU.

\begin{table*}[t]
\centering
\caption{Comparative evaluation of various visuomotor control strategies}
\label{tab:comparison}
\resizebox{\textwidth}{!}{\begin{tabular}{|c|c|c|c|c|c|c|}
\hline
\textbf{Techniques}           & \textbf{Requires} & \textbf{General}  & \textbf{Expert}         & \textbf{Computationally} & \textbf{Safety}     & \textbf{Completeness} \\
                              & \textbf{Model}    & \textbf{Dynamics} & \textbf{Demonstrations} & \textbf{Expensive}       & \textbf{Guarantees} & \textbf{Guarantees}   \\ \hline
Imitation Learning \cite{Osa_2018, pan2019agile} & \textbf{\xmark} & \textbf{\cmark} & \cmark         & \textbf{\xmark} & \xmark           & \xmark \\ \hline
NeRF CBF  \cite{10161482}   & \cmark         & \xmark           & \textbf{\xmark} & \cmark         & \textbf{\cmark} & \xmark \\ \hline
Vision CBF\cite{10160805}    & \textbf{\xmark} & \xmark           & \textbf{\xmark} & \cmark         & \textbf{\cmark} & \xmark \\ \hline
Neural CBF \cite{harms2024neuralcontrolbarrierfunctions}       & \cmark         & \xmark           & \textbf{\xmark} & \textbf{\xmark} & \textbf{\cmark} & \xmark \\ \hline
Latent CBF \cite{kumar2024latentcbf}    & \textbf{\xmark} & \xmark           & \textbf{\xmark} & \textbf{\xmark} & \textbf{\cmark} & \xmark \\ \hline
Semi-Supervised Policy (Ours) & \textbf{\xmark}       & \textbf{\cmark}      & \textbf{\xmark}             & \textbf{\xmark}              & \textbf{\cmark}        & \textbf{\cmark}          \\ \hline
\end{tabular}}
\end{table*}

\subsection{Inverted Pendulum}
In our first case study, we analyze an inverted pendulum system characterized by the state vector $x = [\Theta, \dot{\Theta}] \in \left[-\pi, \pi\right] \times \left[-3.5, 3.5\right]$, where $\Theta$ represents the angular position, and $\dot{\Theta}$ denotes the angular velocity. The visuomotor observation $\mathcal{O}_{t}$ consists of a stack of two RGB frames, each with dimensions of 64$\times$64 pixels. The control input to the system is the applied torque $\tau \in [-10, 10]$ Nm, as illustrated in Fig. \ref{fig:sys_pend}. The discrete-time dynamics governing this system are given by:
\begin{equation}
    \left[\begin{array}{l}
    \Theta_{t+1} \\
    \dot{\Theta}_{t+1}
    \end{array}\right] =
    \left[\begin{array}{l}
    \Theta_{t} \\
    \dot{\Theta}_{t}
    \end{array}\right] + \left(\left[\begin{array}{c}
    \dot{\Theta}_{t} \\
    \frac{g}{l}\sin(\Theta_t)
    \end{array}\right]+\left[\begin{array}{c}
    0 \\
    \frac{1}{m l^{2}} 
    \end{array}\right] u_t\right) dt, \\
    \mathcal{O}_{t} = Img(x_{t}, x_{t-1}),
\end{equation}
where $m$ and $l$ represent the mass and length of the pendulum, respectively, and $Img$ represents the stack of RGB frames at two consecutive time steps.
The datasets, $\mathcal{S}, \mathcal{U},$ and $\mathcal{D}$, are sampled as follows:
\begin{equation}
    \begin{aligned}
    \mathcal{S} &= \{ \mathcal{O}_t |  x_t \in \left[-\frac{\pi}{12}, \frac{\pi}{12}\right]\times \left[-0.25, 0.25\right] \} \\
    \mathcal{U} &= \{\mathcal{O}_t| x_t \in \left[-\pi, \pi\right]\times \left[-3.5, 3.5\right]  \backslash \left[-\frac{\pi}{2}, \frac{\pi}{2}\right] \times \left[-1.5, 1.5\right] \} 
    \\
    \mathcal{D} &= \{\mathcal{O}_t| x_t \in \left[-\pi, \pi\right] \times \left[-3.5, 3.5\right]\}. 
    \end{aligned}
\end{equation}
The barrier network used in this case is Lipschitz bounded with $L_B = 2$, and the latent dimension size is set to 2. A user-defined policy, $\pi_{user}$, is set to 0 to enable the training of a robust controller. 

Visualizations of the trained barrier function are presented in both latent space with sample points (Fig. \ref{fig:plot_latent_pend}) and in $\Theta-\dot\Theta$ plane (Fig. \ref{fig:plot_real_pend}). 
These visualizations demonstrate the successful separation of the safe region from the unsafe region. As shown in Fig. \ref{fig:sspl_pend}, the trajectories generated by the learned policy $\pi_{\theta}$ start in the unsafe region and successfully reach the safe region, validating our approach.

\subsection{Obstacle Avoidance on Autonomous Ground Vehicle}

\begin{figure}[t]
\begin{subfigure}{.47\linewidth}
    \centering
    \includegraphics[width=\linewidth]{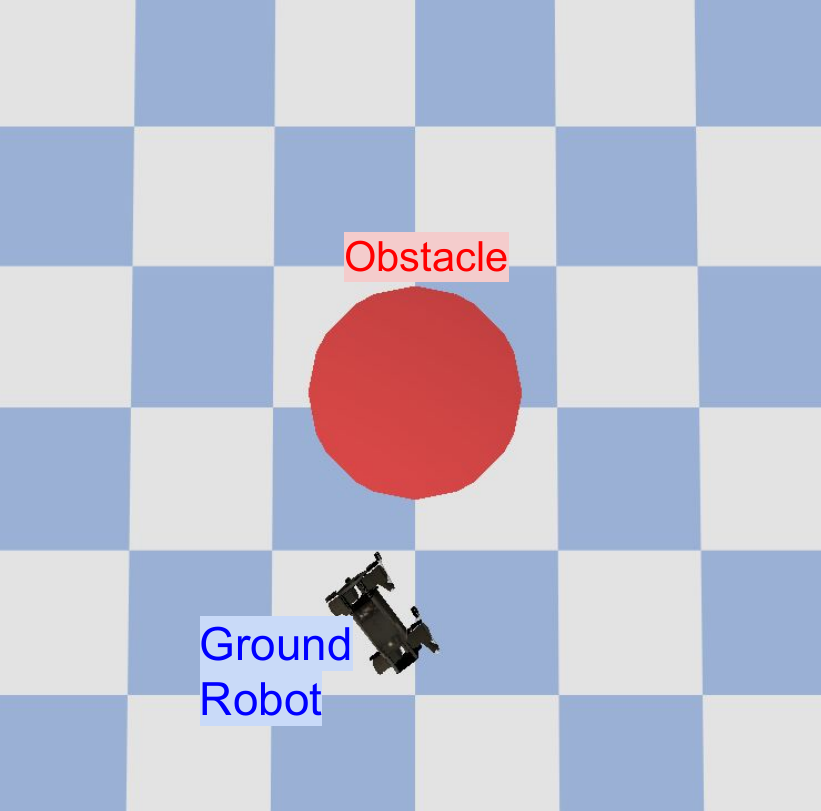}
    \subcaption{}
    \label{fig:sspl_gr}
\end{subfigure}%
\hfill
\begin{subfigure}{.47\linewidth}
    \centering
    \includegraphics[width=\linewidth]{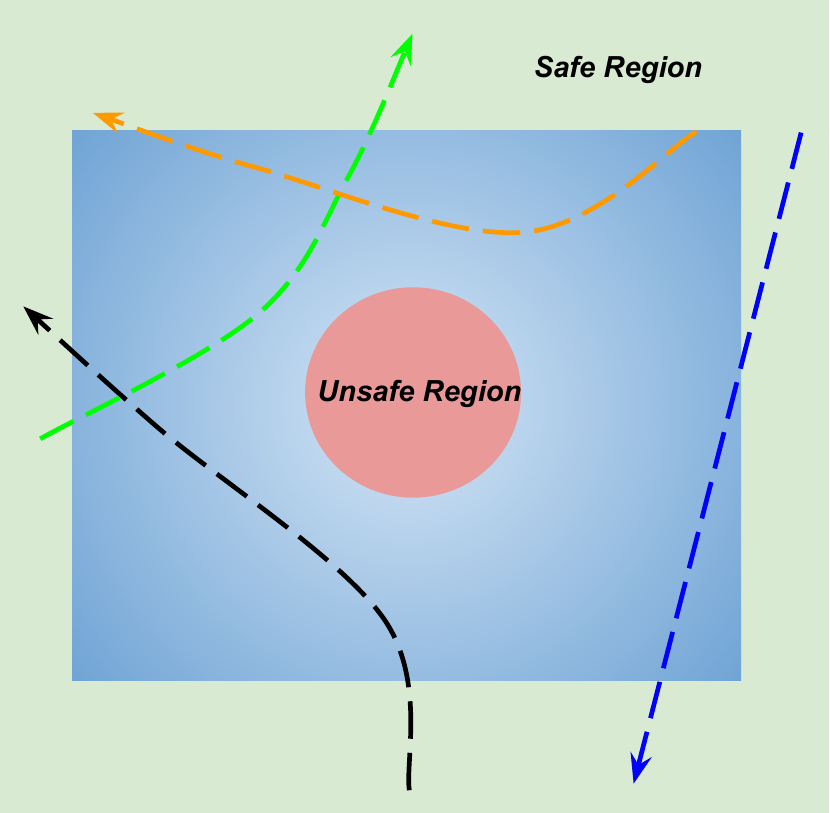}
    \subcaption{}
    \label{fig:traj_sspl_gr}
\end{subfigure}%

\caption{Presents the results on the velocity-driven quadruped model on PyBullet. 
\textbf{Fig. \ref{fig:sspl_gr}} shows the top view of the $x-y$ plane, which represents the visuomotor input of the system.
\textbf{Fig. \ref{fig:traj_sspl_gr}} shows the x-y plane where the light green area represents the safe region (used for sampling safe states, $\mathcal{S}$), and the light red area denotes the unsafe region (used for sampling unsafe states, $\mathcal{U}$). The figure also shows multiple trajectories initiating inside the safe set following the policy $\pi_{\theta}$, synthesized using the proposed framework.
}
\end{figure}
In our next case study, we analyze an obstacle avoidance problem for an autonomous ground robot, modeled here as a reduced-order representation of a quadruped robot \cite{C3BF-Legged, tayal2024polygonal, molnar2021model}. The state vector is defined as $[x_t, y_t, \Theta_t]^T \in [-2, 2]^2 \times [-\pi, \pi]$, where $(x_t, y_t)$ denotes the robot's position and $\Theta_t$ represents its orientation. The robot moves at a constant speed $v = 1$, while the visuomotor observation $\mathcal{O}_{t}$ consists of a stack of two 128$\times$128 RGB frames. The control input $u$ corresponds to the yaw rate $\omega$.

It follows the discrete-time dynamics given by \cite{dubins1957curves}:
\begin{equation}
    \left[\begin{array}{l}
    x_{t+1} \\
    y_{t+1} \\
    \Theta_{t+1}
    \end{array}\right]=
    \left[\begin{array}{l}
    x_{t} \\
    y_{t} \\
    \Theta_{t}
    \end{array}\right]
    +    
    \left(\left[\begin{array}{c}
    v\cos\Theta_t \\
    v\sin\Theta_t \\
    0
    \end{array}\right]+\left[\begin{array}{c}
    0 \\
    0 \\
    1 
    \end{array}\right] u_t\right) dt, \\
    \mathcal{O}_{t} = Img(x_{t}, x_{t-1}),
\end{equation}
where $Img$ represents the stack of RGB frames at two consecutive time steps. Fig \ref{fig:sspl_gr} shows a sample input image at a particular time instant.
The datasets, $\mathcal{S}, \mathcal{U}, \mathcal{D}$, are sampled as follows:
\begin{equation}
    \begin{aligned}
    \mathcal{S} &= \{ \mathcal{O}_t |  x_t \in [-2,2]^2 \times [-\pi,\pi] \backslash [-1.5, 1.5]^{2} \times [-\pi,\pi] \} \\
    \mathcal{U} &= \{\mathcal{O}_t| x_t \in [-0.7, 0.7]^{2} \times [-\pi,\pi] \} 
    \\
    \mathcal{D} &= \{\mathcal{O}_t| x_t \in [-2,2]^2 \times [-\pi,\pi]\} 
    \end{aligned}
\end{equation}

The barrier network used in this case is Lipschitz bounded with $L_B = 1.5$, and the latent dimension size is set to 4.

We assess the performance of the proposed safe policy within the Quadruped environment using the PyBullet simulation framework \cite{coumans2019}. The quadruped robot is equipped with a low-level controller, such as Convex MPC, which facilitates precise velocity tracking along the $x$, $y$, and yaw axes.

Figure \ref{fig:traj_sspl_gr} illustrates the trajectories generated by the learned policy $\pi_{\theta}$, which successfully avoids the obstacle and reaches the safe region, thereby validating our approach.

\subsection{Comparative Study with other state-of-the-art techniques}
Table \ref{tab:comparison} provides a comparative analysis of different visuomotor control techniques, assessed on multiple criteria. These criteria include the need for a system model, the capability to manage general dynamics, the reliance on expert demonstrations, computational complexity, safety assurances, and whether these assurances extend across the entire state space (completeness guarantees). 

It is evident that the proposed framework ensures safety guarantees for systems with general dynamics without relying on a model or expert demonstrations. Moreover, it is the only framework that offers completeness guarantees for the learned barrier functions, thereby ensuring that the policy is provably safe.

\section{Conclusion}
\label{section: Conclusions}
We introduced a semi-supervised framework that synthesizes a Control Barrier Certificate (CBC) and a safe policy from visuomotor observations. The learned CBC inherently satisfies completeness guarantees, eliminating the need for post hoc verification. Our Safety-based Latent Dynamics (SaLaD) model efficiently distinguishes safe from unsafe regions. Moreover, we derived a consistency condition to address non-zero dynamics, ensuring the policy meets CBC conditions. Experiments on an inverted pendulum and an autonomous mobile robot confirm that our framework generates policies that adhere to safety constraints. 

\textbf{Limitations and Future Work}: The framework guarantees safety upon convergence, but the convergence rate is not addressed. Future work will focus on improving and analyzing convergence rates, as well as extending the framework to more complex systems.

\label{section: References}
\bibliographystyle{IEEEtran}
\bibliography{ref.bib}


\end{document}